\documentclass[10pt,twocolumn,letterpaper]{article}

 \usepackage{cvpr}              

\usepackage[dvipsnames]{xcolor}

\definecolor{cvprblue}{rgb}{0.21,0.49,0.74}
\usepackage[pagebackref,breaklinks,colorlinks,citecolor=cvprblue]{hyperref}

\title{Enhancing Weakly-Supervised Object Detection on Static Images through (Hallucinated) Motion }

\begin{document}
\newcommand*{\affaddr}[1]{#1} 
\newcommand*{\affmark}[1][*]{\textsuperscript{#1}}

\author{%
Cagri Gungor and Adriana Kovashka\\
\affaddr{University of Pittsburgh}\\
{\tt\small cagri.gungor@pitt.edu}, {\tt\small kovashka@cs.pitt.edu}\\
}
\maketitle
\begin{abstract}
While motion has garnered attention in various tasks, its potential as a modality for weakly-supervised object detection (WSOD) in static images remains unexplored. Our study introduces an approach to enhance WSOD methods by integrating motion information. This method involves leveraging hallucinated motion from static images to improve WSOD on image datasets, utilizing a Siamese network for enhanced representation learning with motion, addressing camera motion through motion normalization, and selectively training images based on object motion. Experimental validation on the COCO and YouTube-BB datasets demonstrates improvements over a state-of-the-art method.\end{abstract}    
\section{Introduction}
\label{sec:intro}

Weakly-supervised object detection (WSOD) faces challenges in determining which instance carry training image-level labels, with traditional methods \cite{bilen2016weakly,ren2020instance,sui2022salvage,unal2022learning} relying primarily on appearance information in RGB images. While static appearance details are an appropriate foundation, their limitation becomes evident in dynamic scenarios. Incorporating a modality capturing motion and temporal dynamics, provides distinct and complementary information to appearance. This integration presents an opportunity to enhance the localization of objects, particularly in scenarios with dynamic behaviors, by leveraging the temporal insights offered by motion information. For example, the motion of a car might indicate its trajectory, speed, or interaction with other objects, providing crucial context beyond what static appearance can offer. 

Our ultimate goal is to enhance object detection performance on static images in the COCO dataset by leveraging motion. As a preliminary step, we present a proof of concept on the YouTube-BB video dataset, where real motion exists between frames. Our methodology introduces a Siamese WSOD network with contrastive learning, integrating motion to enhance representation learning during training. We employ motion normalization to reduce camera motion in the video dataset, ensuring more reliable motion information. Moreover, we strategically select images on training set based on object motion to obtain a training set that includes images with heightened and meaningful motion, amplifying the potential influence of motion while minimizing noise from low-quality motion and images with restricted motion. 
Lastly, we extract hallucinated motion from static images, supporting our ultimate object that motion can enhance object detection even in static images.


\begin{figure}[t]
    \centering
    \includegraphics[width=1\linewidth]{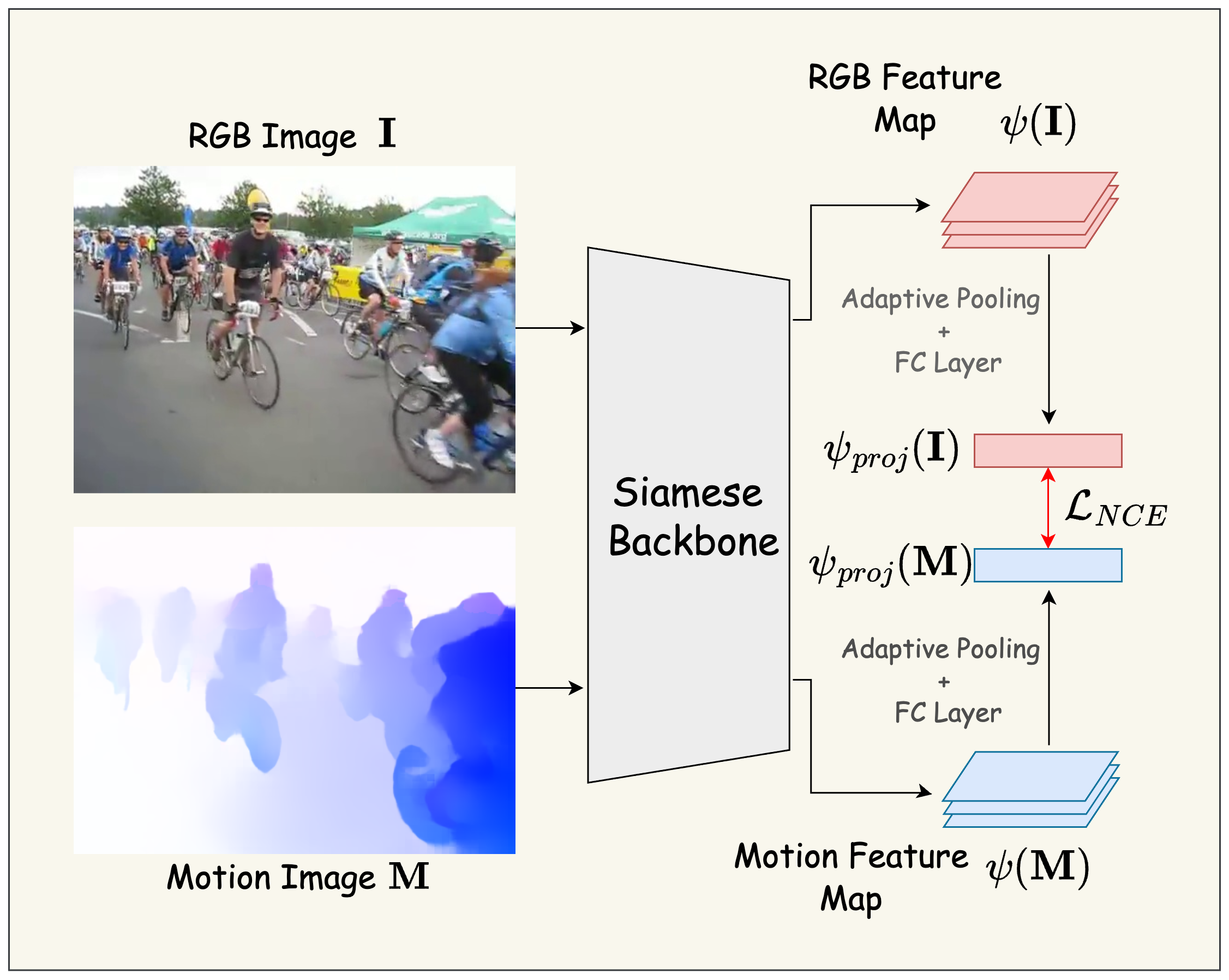}
    \caption{This figure illustrates the design of a Siamese WSOD network and contrastive learning by leveraging the motion modality to improve representation learning.}
 
    \label{fig:main}
\end{figure}

In prior work, W-RPN \cite{singh2019you} improves proposal quality in WSOD through motion by training an RPN before WSOD training. Pathak et al. \cite{pathak2017learning} suggest using unsupervised motion-based grouping for object segmentation, aiming to learn improved features.
In contrast, our approach directly utilizes motion during WSOD training, demonstrating that hallucinated motion enhance detection performance in static images.
\section{Approach}
\label{sec:method}

\begin{figure}[t]
    \centering
    \includegraphics[width=1\linewidth]{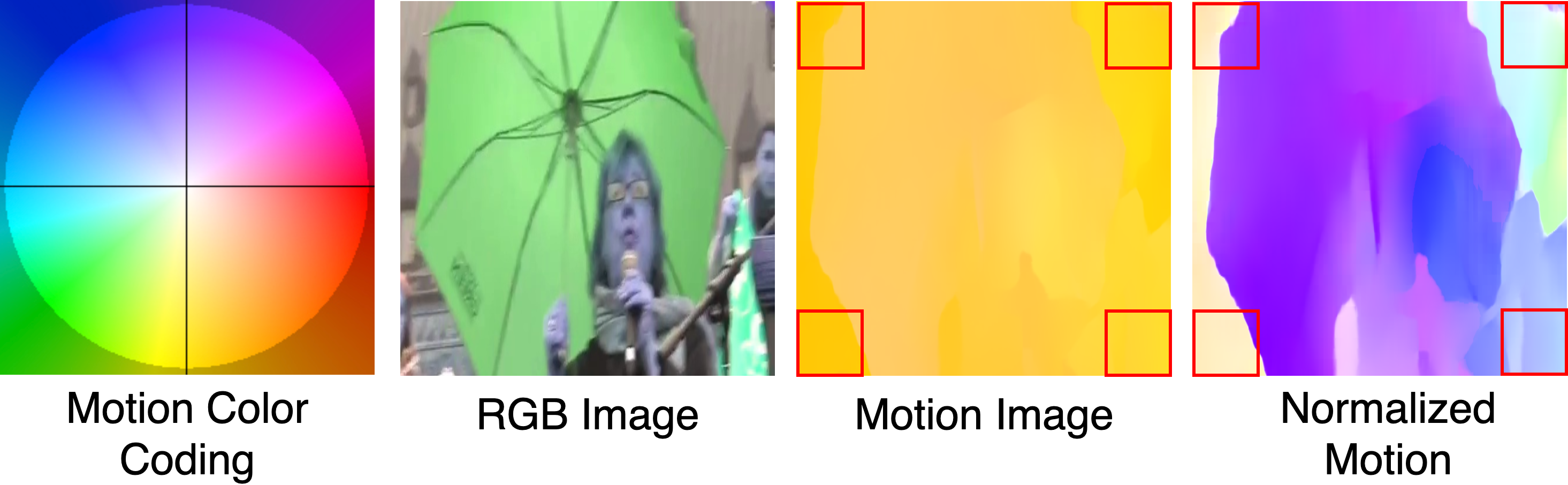}
    \caption{{Visualization of the motion normalization approach.   }}
    \label{fig:normalization}
\end{figure}
\subsection{The Siamese WSOD Network}
\label{sec:siamese_module}

\textbf{WSOD.}
Following \cite{ren2020instance}, the RGB image $\mathbf{I}$ undergoes RoI pooling for each of $R$ visual proposals $v_i$ to generate a fixed-length feature vector $\phi(v_i)$. Detection and classification scores are computed for each proposal and category using parallel fully-connected layers. The ground-truth class label for each image is represented as $y_c \in {0, 1}$, where $c \in {1, . . . , C}$ and $C$ is the total number of object categories.

The detection and classification scores are computed:
\begin{equation}
v^{det}_{i,c} = w^{det\intercal}_{c} \phi(v_i) + b^{det}_{c}, \quad v^{cls}_{i,c} = w^{cls\intercal}_{c} \phi(v_i) + b^{cls}_{c}
\label{eq:1}
\end{equation}
where $w$ and $b$ represent the weights and biases.
These scores are then transformed into probabilities where $p^{cls}_{i,c}$ denotes the probability of class $c$ being present in proposal $v_i$, while $p^{det}_{i,c}$ signifies the probability that $v_i$ is crucial for predicting the image-level label $y_c$.
\begin{equation}
   p^{det}_{i,c} = \dfrac{exp(v^{det}_{i,c})} {\sum_{k=1}^{R} exp(v^{det}_{k,c})}, \quad p^{cls}_{i,c} = \dfrac{exp(v^{cls}_{i,c})} {\sum_{k=1}^{C} exp(v^{cls}_{i,k})}
   \label{eq:7}
\end{equation}


Ultimately, image-level predictions $\hat{p}_c$ are calculated and used for training in the absence of region-level labels:
\begin{equation}
  \hat{p}_c = \sigma\left(\sum_{i=1}^{R}
  p^{det}_{i,c} p^{cls}_{i,c}
  \right)
  \label{eq:confidence}
\end{equation}
\begin{equation}
    \mathcal{L}_{mil} = -\sum_{c=1}^{C}\left[ y_c\log\hat{p}_c + (1-y_c)\log(1-\hat{p}_c) \right]
    \label{eq:visual_loss}
\end{equation}

\textbf{Obtaining motion images.} FlowNet 2.0 \cite{ilg2017flownet} is employed to compute the optical flow between the image and the successive frame in the video. The resulting optical flow is referred to as GT motion images. Additionally, we leverage Im2Flow \cite{gao2018im2flow} to hallucinate motion from static images, thereby obtaining the motion modality in the image dataset. The motion images contain 2 channels, representing the horizontal and vertical components of the optical flow. These 2-channel motion images are converted into RGB images by applying color coding \cite{chantas2014variational}, as visualized in Fig.~\ref{fig:normalization}. Each distinct color in the resulting images corresponds to a different direction of motion. Additionally, darker colors signify a higher magnitude of motion. This transformation enables them to utilize the same siamese backbone as the RGB images.

\textbf{Siamese design.} In our approach to weakly-supervised object detection, we integrate motion information through a Siamese network \cite{gungor2024boosting, fu2021siamese, song2021exploiting, meyer2020improving} employing contrastive learning during training. This strategy allows the utilization of a pre-trained RGB backbone to extract features from both RGB and motion images, without introducing additional complexity to the model. Representation learning is enhanced by introducing a contrastive loss between RGB and motion features. The design of our Siamese network enables the utilization of only RGB images during inference, ensuring no extra overhead on inference time.

As shown in Fig.~\ref{fig:main}, we extract feature maps from the RGB image $\mathbf{I}$ and the motion image $\mathbf{M}$ using a pre-trained backbone. These feature maps are then passed through adaptive pooling and a Siamese fully connected layer, resulting in projected feature vectors $\psi_{proj}(\mathbf{I})$ and $\psi_{proj}(\mathbf{M})$.

\textbf{Contrastive learning.}
The RGB and motion feature vectors, denoted as $\psi_{proj}(\mathbf{I})$ and $\psi_{proj}(\mathbf{M})$ respectively, undergo L2-normalization. Following this normalization, we compute their cosine similarity:
\begin{equation}
    S(\mathbf{I},\mathbf{M}) = 	\langle \psi_{proj}(\mathbf{I}), \psi_{proj}(\mathbf{M}) \rangle / \rho
    \label{eq:b}
\end{equation} where $\rho$, is a learnable temperature parameter.

Given RGB and motion image pairs $(\mathbf{I}, \mathbf{M}) \in \mathcal{B}$, where $\mathcal{B}$ represents an RGB-motion pair batch, we employ noise contrastive estimation (NCE) \cite{gutmann2010noise}. The NCE loss contrasts an RGB image with negative motion images to assess the similarity between the RGB image and its paired motion image, relative to others in the batch:

\begin{equation}
\resizebox{\columnwidth}{!}{$
    \mathcal{L}_{{M} \rightarrow {I}} = -\dfrac{1}{|\mathcal{B}|} \sum_{(\mathbf{I},\mathbf{M}) \in \mathcal{B}} log \dfrac{exp(S(\mathbf{I},\mathbf{M}))}{exp(S(\mathbf{I},\mathbf{M})) + \sum_{(\mathbf{I}',\mathbf{M}') \in \mathcal{B}} exp(S(\mathbf{I},\mathbf{M}'))}$}
\end{equation}

The second component of the NCE loss, denoted as $\mathcal{L}_{{I} \rightarrow {M}}$, is similarly formulated to contrast a motion image with negative RGB image samples. The average of these two components forms the complete NCE loss:

\begin{equation}
    \mathcal{L}_{NCE} = (\mathcal{L}_{{M} \rightarrow {I}} + \mathcal{L}_{{I} \rightarrow {M}}) / 2
\end{equation}

\subsection{Normalization to Tackle Camera Motion}
\label{sec:normalization}

While object motion yields valuable insights into object characteristics and positioning, the presence of camera motion introduces interference, creating noise and hindering the accurate extraction of object motion. Due to the ubiquity of camera motion across images, we conceive an approach to extract camera motion, aiming to enhance the accuracy of object motion extraction.

We make an assumption that the motion values in the corners of an image approximate the background motion. Consequently, we calculate the background motion by taking into account the motion observed in these corners. We first compute per-corner motion values (scalars) by averaging per-pixel values. 
We then cluster the four corner values into two clusters, and drop any singleton clusters (one corner), to reduce the effect of a corner having any relative mismatch in motion values among the other corners, such as when a part of an object is situated on a corner. Further, we take the average motion of corners to obtain the approximated background motion. Because the background motion is caused by camera motion, we subtract the background motion value from the image to obtain a more accurate object motion estimate.

In Fig.~\ref{fig:normalization}, the motion image exhibits a consistent yellow tone across the image, indicating downward camera motion according to the color coding scheme. Our method focuses on the image corners outlined with red rectangles to estimate camera motion. Notably, the exclusion of the right upper corner, affected by object (rather than camera) motion, contributes to the improved accuracy of camera motion approximation. Following the normalization process to tackle camera motion, the depiction of the object ``umbrella" illustrates a more accurate representation of its motion.

\begin{figure}[t]
    \centering
    \includegraphics[width=1\linewidth]{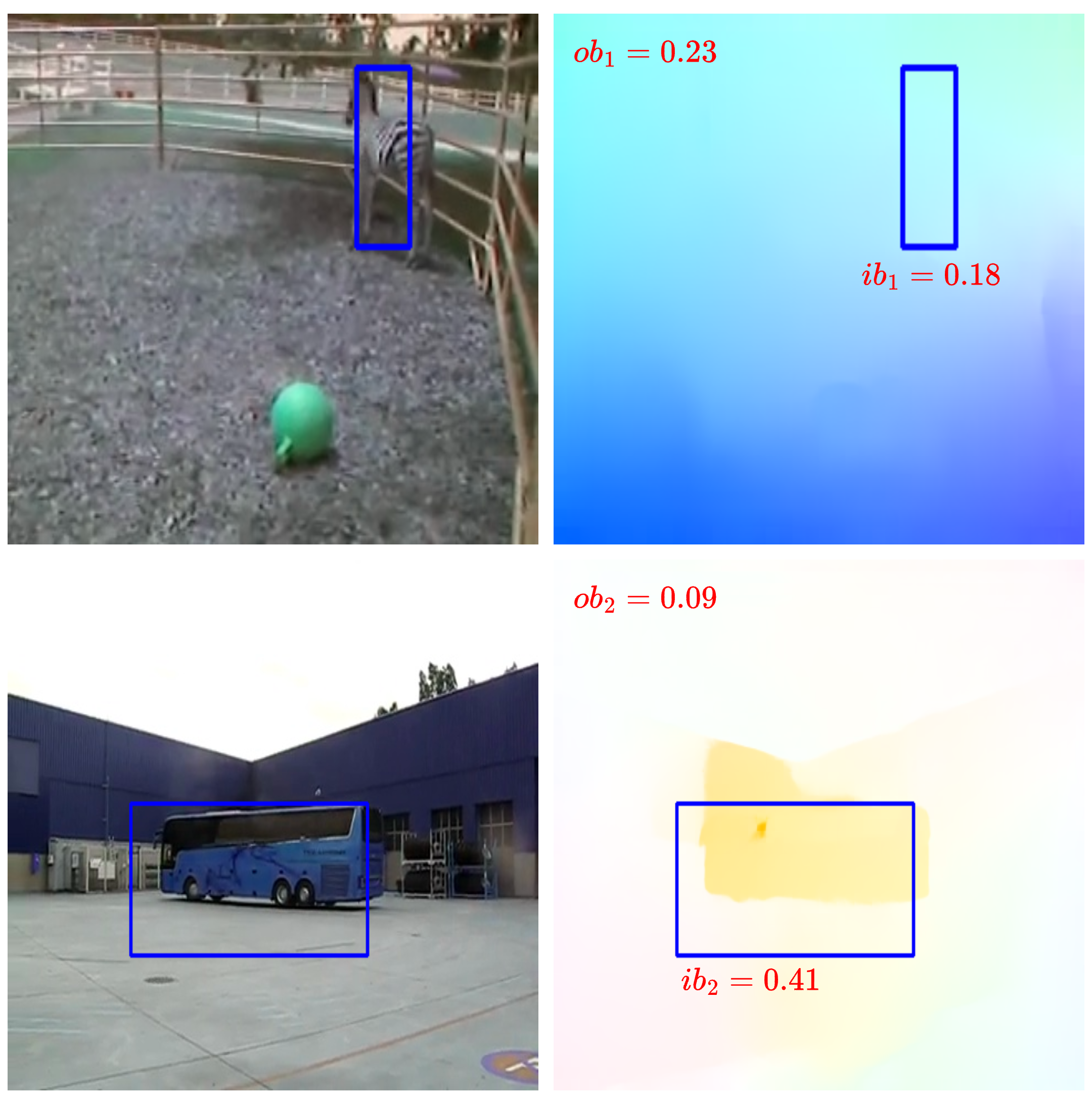}
    \caption{{Visualization of image selection based on motion. The top row image is not selected due to the absence of object motion, while the second row image is selected for its noticeable motion.}}
    \label{fig:selection}
\end{figure}

\vspace{-0mm}

\subsection{Motion-Driven Training Image Selection}
\label{sec:selection}

We choose images in the training set depending on whether there is significant object motion. 
Following the selection process, the training set include images with more pronounced and meaningful motion, amplifying the potential influence of motion while minimizing noise from low-quality motion and images with limited motion and enhancing detection performance. We employ the selection only on training set as we do not use motion during inference. 

The process by which we determine if there is significant object motion involves comparing object and background motion (described next). Thus, it requires that we first estimate which bounding boxes containing objects. 
Employing ground truth bounding boxes (even just at training time) would violate the WSOD setting, which relies solely on class labels for supervision. To adhere to this setting, we leverage a baseline WSOD method, which serves as a foundation for improvement, generating bounding box predictions for objects in the training set. 

Let $ib_i$ denote the average magnitude motion value inside the bounding box prediction, where $i \in {1, . . . , N}$ and $N$ is the total number of images in the training set. On the other hand, $ob_i$ represents the average magnitude motion value outside the bounding box prediction. The magnitude motion values are normalized between 0 and 1. Our image selection criterion involves choosing an image if $ib_i$ surpasses the minimum motion threshold $m = 0.2$ and if the ratio between $ib_i$ and $ob_i$ exceeds the minimum motion ratio threshold $d = 1.5$ ($m$ and $d$ chosen by by subjectively evaluating multiple cases to determine the best fit.). Following this criterion, the variable $s_i$ signifies whether an image is selected or not:

\begin{equation}
    s_i= 
\begin{cases}
    1,& \text{if } ib_i > m \text{ and } ib_i / ob_i > d\\
    0,              & \text{otherwise}
\end{cases}
\end{equation}

This ensures objects in training images exhibit sufficient motion overall and compared to the background. Note the YouTube-BB dataset where we apply selection usually features a single object.

In Fig.~\ref{fig:selection}, the image in the top row features a horse, but it is identified as a non-moving object with a magnitude of motion inside box $ib_1 = 0.18$, while the magnitude of motion outside of the box $ob_1 = 0.23$. Thus, our method excludes this image from the WSOD training set due to the weak motion. In contrast, the bus in the image below has substantial motion, as evidenced by a magnitude of motion inside  $ib_2$ at $0.41$, compared to a magnitude of motion outside  $ob_2$ at $0.09$. Therefore, this image is selected for training to investigate the impact of motion on the WSOD task.
\section{Experiments}
\label{sec:experiments}

We test our method on top of one of the SOTA weakly-supervised detection methods MIST \cite{ren2020instance} which operates in an RGB-only setting, and verify the contributions of our approach which are Siamese WSOD network with motion (Sec.~\ref{sec:siamese_module}), motion normalization (Sec.~\ref{sec:normalization}) and training image selection based on motion (Sec.~\ref{sec:selection}). 

\subsection{Datasets}

\textbf{Common Objects in Context (COCO)} \cite{lin2014microsoft} is a large-scale image dataset for object detection, segmentation, and captioning.  We focus on 18 moving object classes (bird, cat, train, boat, umbrella, motorcycle, elephant, bus, cow, car, horse, bicycle, truck, airplane, giraffe, zebra, dog, bear). 
We extract hallucinated motion from images to show motion helps object detection of even static images. The training set comprises approximately 32k images, while the test set consists of around 2k images.

\textbf{YouTube-BB} \cite{real2017youtube} is a  video dataset, from which we curate a subset image dataset containing the same classes as those in the COCO dataset by randomly sampling images from videos. GT motion is derived by computing optical flow between each image and its consecutive frame in the video. The training set initially contains about 38k images. Following motion-driven image selection, the training set is refined to approximately 27k images. We employ the same test set utilized in the COCO dataset for our evaluations. The video dataset is preliminary step and proof of concept to reach our ultimate goal (hallucinated motion).

\subsection{Comparison against MIST and ablations}
\begin{table}[t]
\centering
\resizebox{\columnwidth}{!}{
\begin{tabular}{ lccc}
\toprule
 & \multicolumn{3}{c}{Avg. Precision, IoU}  \\
 \cmidrule{2-4} 
YouTube-BB & 0.5:0.95 & 0.5  & 0.75  \\

\midrule

\textsc{RGB \cite{ren2020instance}} & 2.8 & 9.8 & 1.0 \\
+ GT Motion & 3.1 (\textcolor{green}{+10\%}) &10.2 (\textcolor{green}{+4\%}) &{1.1} (\textcolor{green}{+10\%}) \\
+ GT Normalized Motion & \bf{3.1} (\textcolor{green}{+10\%})&\bf{10.4} (\textcolor{green}{+6\%})&\bf{1.2} (\textcolor{green}{+20\%})\\
+ Hallucinated Motion & {2.7} (\textcolor{red}{-3.0\%}) &{9.3} (\textcolor{red}{-5.0\%}) &1.0 (\textcolor{gray}{-})\\

\midrule
\midrule
YouTube-BB /w selection   \\
\midrule

\textsc{RGB \cite{ren2020instance}} & 2.6 & 9.2 & 0.8 \\
+ GT Motion & 3.0 (\textcolor{green}{+15\%}) &10.6 (\textcolor{green}{+15\%}) &{0.9} (\textcolor{green}{+13\%})\\
+ GT Normalized Motion & \bf{3.0} (\textcolor{green}{+15\%}) &\bf{10.8} (\textcolor{green}{+17\%}) &\bf{1.0} (\textcolor{green}{+25\%}) \\
+ Hallucinated Motion & {2.9} (\textcolor{green}{+12\%}) &{10.1}(\textcolor{green}{+10\%}) &0.9 (\textcolor{green}{+13\%})\\

\midrule
\midrule
COCO   \\
\midrule
\textsc{RGB \cite{ren2020instance}} & 11.8 & 27.5 & 8.3 \\
+ Hallucinated Motion & \bf{12.1} (\textcolor{green}{+3.0\%})&\bf{27.6} (\textcolor{green}{+0.4\%})&\bf{8.4} (\textcolor{green}{+1.2\%})\\

\bottomrule
\hline \\
\end{tabular}
}
\caption{This table compares our proposed methods and the baseline  MIST \cite{ren2020instance} which uses RGB only. The evaluation is conducted on YouTube-BB, YouTube-BB with motion-driven selection, and COCO datasets. Performance improvements are highlighted in green, indicating the percentage increase, 
while performance declines are marked in red.
The best performer per column is in \textbf{bold}.} 
\label{Table:1}
\end{table}

We enhance the MIST \cite{ren2020instance} baseline by implementing a Siamese network with contrastive learning to incorporate motion modality and improve representation learning on the backbone. The integration of GT motion results in notable improvements, showcasing a 4-10\% and 13-15\% increase in mAP across various thresholds for YouTube-BB and YouTube-BB /w selection datasets, respectively.

Implementing normalization on GT motion images, aimed at mitigating noise introduced by camera motion, yields improved performance. The GT normalized motion exhibits an advancement over the original GT motion, with a 2-10\% and 3-12\% increase in mAP for the YouTube-BB and YouTube-BB with selection datasets, respectively.

To more effectively evaluate the influence of motion on improving detection results, we employ a selective approach in image curation from YouTube-BB, as detailed in Sec.~\ref{sec:selection}, resulting in the YouTube-BB with selection training set. Due to the substantial reduction in the number of images in training set after the selection process, from 38k to 27k, the results in RGB \cite{ren2020instance} are worse in YouTube-BB with selection. However, upon incorporating motion in both GT Motion and GT Normalized Motion settings, we observe higher performance improvement over RGB setting, ranging from 13-25\%.
This improvement surpasses the enhancement observed in the YouTube-BB dataset, which ranged from 4-20\%. Thus normalization and selection allow us to observe the influence of motion and empower the proof of concept that motion boosts object detection.

After demonstrating the impact of GT and hallucinated motion on frames from a video dataset, we provide evidence supporting our ultimate goal that hallucinated motion using Im2Flow \cite{gao2018im2flow} enhances object detection even in static images. The quality of the hallucination is notably compromised in scenarios involving complex backgrounds, non-moving entities, and small objects. The integration of hallucinated motion with RGB images in the YouTube-BB dataset results in a decrease in performance due to the poor quality of the generated motion. After selecting training images based on the amount of motion in YouTube-BB with selection training set, more reliable hallucinated motion is obtained on the average of training set. Thus we observe a noteworthy improvement of 10-13\% in the YouTube-BB with selection dataset when leveraging hallucinated motion.

Ultimately, we apply hallucinated motion to the COCO image dataset where GT motion is unavailable, yielding a performance improvement ranging from 0.4-3\%. This underscores the capability of motion to enhance the accuracy of object detection in static images.
\section{Conclusion}
\label{sec:conclusion}

Our method enhances WSOD by integrating motion information. The utilization of GT motion in the video dataset serves as a proof of concept, demonstrating that motion is a complementary modality to vision, improving object detection. The inclusion of hallucinated motion supports our ultimate goal, indicating that object detection in static images can be enhanced through motion.
{
    \small
    \bibliographystyle{ieeenat_fullname}
    \bibliography{main}
}


\end{document}